\documentclass[runningheads]{llncs}
\usepackage{graphicx}
\usepackage{glossaries}
\usepackage[table,xcdraw]{xcolor}
\usepackage{caption}
\usepackage{subcaption}
\usepackage{float}
\usepackage{hyperref}

\begin{document}
\newacronym{ge}{GE}{Grammatical Evolution}
\newacronym{ga}{GA}{Genetic Algorithm}
\newacronym{gp}{GP}{Genetic Programming}
\newacronym{pge}{PGE}{Probabilistic Grammatical Evolution}
\newacronym{sge}{SGE}{Structured Grammatical Evolution}
\newacronym{dsge}{DSGE}{Dynamic Structured Grammatical Evolution}
\newacronym{ea}{EA}{Evolutionary Algorithm}
\newacronym{cfg}{CFG}{Context-Free Grammar}
\newacronym{pcfg}{PCFG}{Probabilistic Context-Free Grammar}
\newacronym{eda}{EDA}{Estimation Distribution Algorithm}
\newacronym{copge}{Co-PGE}{Co-evolutionary Probabilistic Grammatical Evolution}
\newacronym{pmbge}{PMBGE}{Probabilistic Model Building Grammatical Evolution}
\newacronym{cdt}{CDT}{Conditional Dependency Tree}
\newacronym{nt}{NT}{Non-terminal}
\newacronym{pige}{$\pi$GE}{Position Independent Grammatical Evolution}
\newacronym{cfggp}{CFG-GP}{Context-Free Grammar Genetic Programming}
\newacronym{rrse}{RRSE}{Root Relative Squared Error}
\newacronym{ec}{EC}{Evolutionary Computation}
\newacronym{adf}{ADF}{Automatically Defined Functions}
\newacronym{psge}{PSGE}{Probabilistic Structured Grammatical Evolution}
\newacronym{tag}{TAG}{Tree-Adjunct Grammar}
\newacronym{tage}{TAGE}{Tree-Adjunct Grammatical Evolution}
\newacronym{pigrow}{PI Grow}{Position Independent Grow}
\newacronym{ptc2}{PTC2}{Probabilistic Tree Creation 2}
\newacronym{mi}{MI}{Mutation Innovation}
\newacronym{ci}{CI}{Crossover Innovation}
\newacronym{ai}{AI}{Artificial Intelligence}
\newacronym{copsge}{Co-PSGE}{Co-evolutionary Probabilistic Structured Grammatical Evolution}
\newacronym{mgga}{mGGA}{meta-Grammar Genetic Algorithm}
\newacronym{ge2}{(GE)$^2$}{Grammatical Evolution by Grammatical Evolution}
\newacronym{es}{ES}{Evolution Strategies}
\newacronym{g3}{G3}{Generalized Generation Gap}
\newacronym{apmga}{APmGA}{Adaptive Mutation Probability Genetic Algorithm}
\newacronym{frm}{FRM}{Fitness Reactive Mutation}
\newacronym{mbf}{MBF}{Mean Best Fitness}
\newacronym{chavela}{CHAVELA}{class of hybrid adaptive evolutionary algorithms}
\newacronym{haea}{HAEA}{Hybrid Adaptive Evolutionary Algorithm}
\newacronym{amga}{AMGA}{Adaptive and Modular Genetic Algorithm}
\newacronym{afm}{AFM}{Adaptive Facilitated Mutation}
\newacronym{fm}{FM}{Facilitated Mutation}
\title{Context Matters: Adaptive Mutation for Grammars}
\author{Pedro Carvalho\orcidID{0000-0003-3845-4617} \and
Jessica Mégane\orcidID{0000-0001-6697-5423} \and
Nuno Lourenço\orcidID{0000-0002-2154-0642} \and
Penousal Machado\orcidID{0000-0002-6308-6484}
}
\authorrunning{P. Carvalho et al.}
\institute{University of Coimbra, Centre for Informatics and Systems of the University of Coimbra, Department of Informatics Engineering
\email{\{pfcarvalho,jessicac,naml,machado\}@dei.uc.pt}}
\maketitle              
\begin{abstract}
This work proposes Adaptive Facilitated Mutation, a self-adaptive mutation method for Structured Grammatical Evolution (SGE), biologically inspired by the theory of facilitated variation. 
In SGE, the genotype of individuals contains a list for each non-terminal of the grammar that defines the search space. 
In our proposed mutation, each individual contains an array with a different, self-adaptive mutation rate for each non-terminal.
We also propose Function Grouped Grammars, a grammar design procedure to enhance the benefits of the propose mutation.
Experiments were conducted on three symbolic regression benchmarks using Probabilistic Structured Grammatical Evolution (PSGE), a variant of SGE. 
Results show our approach is similar or better when compared with the standard grammar and mutation.

\end{abstract}

\section{Introduction}

Grammar-based \gls{gp} algorithms have been an important tool for the evolution of computer programs since their inception \cite{whigham,Ryan1998,McKay2010}.
The most popular approach is \gls{ge} which is notable for decoupling the genotype and the phenotype, using a grammar to translate a data structure into an executable program.
The representation and variation operators used by \gls{ge} present some known issues, such as low locality and high redundancy. The first means that small changes in the genotype can cause significant changes in the phenotype, and the second means that most modifications do not affect the phenotype. These characteristics result in a bad trade between exploration and exploitation, which makes the algorithm perform similarly to random search \cite{Whigham2015}.

\gls{sge} \cite{Loureno2018} is a variant of \gls{ge} that uses a different representation for the individuals. The genotype comprises several lists, one for each non-terminal in the grammar, and each list contains the indexes of the rules to be expanded.
\gls{sge} shows better performance when compared to \gls{ge}, and other grammar-based approaches \cite{Loureno2017}, but also improved locality and lower redundancy when compared to standard \gls{ge} \cite{Medvet2017,Loureno2016}, in part due to its operators.
This representation allows the recombination operator to be grammar-aware, preserving the list of each non-terminal. On the other hand, the grammar does not inform the mutation operator.
Mutation in \gls{sge} changes the production rule of the non-terminal selected to mutate.
This operator affects all genes with the same frequency, regardless of grammatical context.
Using a static and equal value for all non-terminals fails to consider that not all mutations are equally destructive.

Specific genes can play an essential role in the solution and, when mutated, may completely ruin the phenotype behavior.
On the other hand, some genes may have a tuning role; in this case, a mutation will only result in a minor adjustment to the solution.
Despite these differences, both types of genes are equal in the eyes of mutation.
Biological processes have evolved to prevent this phenomenon. 
Gerhart et al. \cite{gerhart2007} propose that there are core components vital to the individual which remain unchanged for long periods and regulatory genes that combine existing core components and change frequently.
The result is a system that can quickly adapt to new environments through regulatory changes while preserving the core components that ensure individuals are functional.
We can replicate this behavior in grammar-based algorithms using different adaptive mutation probabilities for each non-terminal. 
This approach enables the system to autonomously regulate mutation rates to match the impact of changes to that non-terminal.
Note that this solution is only as effective as the correlation between non-terminals and mutation impact.
It follows that the effectiveness of this mutation is related to grammar-design \cite{Nicolau2018,nicolau2004,dick2022,Hemberg2008PreIP}, as grammars with more rules enable finer tuning of mutation probabilities.
The grouping of productions within each non-terminal is also relevant; separating low and high impact changes into separate rules should improve performance.

In this work, we propose Adaptive Facilitated Mutation, a biologically inspired grammar-aware self-adaptive mutation operator for \gls{sge} and its variants.
Furthermore, we propose "Function Grouped Grammars", a method for grammar design that empirically outperforms grammars commonly used for regression in \gls{ge}.
We compare our approach to standard grammar and mutation and find that, when combined, Function Grouped Grammars and Adaptive Facilitated Mutation are statistically superior or similar to the baseline in three relevant \gls{gp} benchmarks.

The remainder of this work is structured as follows: 
First, section \ref{sec:background} presents the background necessary to understand the work presented. 
Section \ref{sec:mutation_levels} presents the proposed mutation and grammar-design method. 
Section \ref{sec:experimental_setup} details the experimentation setup used, and Section \ref{sec:results} the experimental results regarding performance and analysis of probabilities. 
Section \ref{sec:conc} gathers the main conclusions and provides insights regarding future work.

\section{Background}
\label{sec:background}

\glspl{ea} are optimization algorithms inspired by the biological processes of natural evolution. 
A population of individuals (candidate solutions) evolves over several generations, guided by a fitness function. 
Similar to nature, these individuals are subject to selection, reproduction, and genetic variation.
These algorithms face known issues such as parameter tuning, premature convergence, and lack of diversity. 
Researchers propose novel representations \cite{handbookge,Loureno2018,Megane2022}, selection methods, genetic operators, and parameter selection approaches to address these issues.

\subsection{Grammar-based Genetic Programming}

\gls{gp} is an \gls{ea} that evolves solutions as programs.
Over the years, researchers have proposed many variants of \gls{gp}, and grammar-based approaches gained more popularity as grammars are helpful to set restrictions to the search space \cite{McKay2010}.

\gls{ge} \cite{handbookge} is the most popular grammar-based \gls{gp} methods.
The individuals' genotype is a string/vector of integers that is translated into a phenotype (an executable function) through a grammar. 
The individuals are subject to selection mechanisms, mutation, and crossover in each generation.

This approach is relevant but suffers from high redundancy \cite{Thorhauer2016}, and poor locality \cite{Rothlauf2006}, damaging the efficiency of evolution. Redundancy is measured by analyzing the proportion of effective mutations, and locality studies how well genotypic neighbors correspond to phenotypic neighbors. Standard \gls{ge} showed performance similar to random search \cite{Whigham2015}, which motivated researchers to propose different initialization methods \cite{Nicolau2017}, representations \cite{Loureno2018,Megane2022}, genetic operators, but also to investigate grammar design \cite{Nicolau2018}.

\gls{pige} \cite{ONeill2004} uses a different representation and mapping mechanism that removes the positional dependency that exists in \gls{ge}. Each codon of the genotype contains two values, \textit{nont} that consists of the non-terminal and \textit{rule} that states the rule index to expand. 
This method improves performance compared with \gls{ge} \cite{Fagan2010}. However, another study found that this method also suffers from poor locality \cite{Medvet2017}.

\gls{sge} \cite{Loureno2016,Loureno2018} proposed a new representation for the genotype and variation operators, which resulted in better performance and fewer issues regarding locality and redundancy \cite{Loureno2017,Medvet2017}. 
The genotype comprises a list of integers (one per grammar non-terminal), with each integer corresponding to the index of a production rule. 
This structure allows mutation to occur inside the same non-terminal and crossover to exchange the list of derivation options. 
Another advantage of this proposal is that only valid solutions are allowed.
\gls{sge} imposes a depth limit on solutions.
Once the limit is surpassed, only non-recursive productions are chosen, forcing individuals to consolidate into a valid genotype.

\gls{psge} \cite{Megane2022} is a recent proposal to \gls{sge} that uses a probabilistic grammar, namely a \gls{pcfg}, to bias the search, and where codons of the genotype are floats. 
Each grammar production rule has a probability of being selected. 
These probabilities change based on the frequency of expansion of that rule on the best individual. 
If the rule is not expanded, the probability decreases. 
This proposal performed better or similarly compared to \gls{sge} and outperformed \gls{ge} in all problems. The evolved grammar also provides information about the features more relevant to the problem \cite{Megane2021}.
Another probabilistic grammar approach is \gls{copsge} \cite{Megane2022gecco}. In this method, each individual has a \gls{pcfg}, which may suffer mutation to the probabilities values. This approach also showed similar or better performance than \gls{sge}.

\subsection{Adaptive mutation rate}
Although most works in the literature use a static parameter for mutation and crossover rates, research shows that dynamic parameters may improve the search and introduce more diversity to the population.
Adaptive mutations have been widely proposed in the literature to tackle some of the issues that \glspl{ea} present. 

Self-adaptive Gaussian mutation has been widely used by \gls{es} \cite{Beyer2004EvolutionS} and adapted into \glspl{ea} \cite{Hinterding}. During the evolutionary process, the mutation rate varies, suffering a Gaussian mutation.
This approach achieves better results than standard mutation and performs similarly to \gls{es}.
Teo \cite{Teo2006} studied a self-adaptive Gaussian mutation operator for the \gls{g3} algorithm, and it outperformed the standard algorithm in two of the four problems tested.


Other approaches consider individuals' fitness when adapting the mutation probability. 
Libelli et al. \cite{MarsiliLibelli2000} and Lis \cite{Lis1995} approaches showed better performance than classical \gls{ga}.

\gls{apmga} \cite{Stark2012ANS} dynamically adjusts the mutation probability during the evolutionary process based on the variations of the population entropy between the current and previous generations. 

Salinas et al. \cite{Cruz_Salinas_2017} proposed an \gls{ea} where operators are \gls{gp} trees. 
In each generation, the probability of an operator increases or decreases based on the individual's performance after the operator.

Gomez proposed \gls{haea} \cite{Gmez2004SelfAO} to adapt the operator probabilities during evolution. Each individual encodes its genetic rates. The probabilities by a random value can change according to the fitness of the offspring. This algorithm inspired other proposals that showed that although for some problems there are no significant improvements in performance, the algorithm can obtain similar results without the need to pre-tuning, needing less computational time \cite{Gmez2021,Montero2007,Montero}.

Coelho et al. \cite{Coelho2016} presented a new hybrid self-adaptive algorithm based on \gls{es} guided by neighborhood structures and tested for combinatorial contains mutation probabilities, and the second contains integer values that control the strength of the disturbance. 
The results were similar to the other approaches tested and showed that the adaptive mutation could escape local optima and balance exploration and exploitation.


To our knowledge, there is only one adaptive mutation rate parameter proposal in \gls{ge}.
Fagan et al. \cite{Fagan2012} propose \gls{frm}, an adaptive mutation that increases the mutation rate in case a fitness plateau is reached to diversify the population and decreases when a new optimum is found, using increments/decrements of 0.01. The approach found similar results as the fixing mutation rate.

\subsection{Grammar-design}
It is possible to design grammar to produce syntactically constrained solutions or to incorporate domain knowledge by biasing the grammar. The grammar's design can significantly impact the search of \gls{ge} \cite{nicolau2004,dick2022,Hemberg2008PreIP}.

Miguel Nicolau \cite{nicolau2004} proposes a method to reduce the number of non-terminals of the grammar. 
The authors compare standard \gls{ge} with a standard and a reduced grammar and showed an empirical increase in performance.
This work motivated a study with different types of grammars.
This study shows that recursion-balanced grammar could also improve performance \cite{Nicolau2018}. 

A recent work by Dick et al. \cite{dick2022} showed that \gls{ge} is more sensitive to grammar design than \gls{cfggp}.
The results suggest that \gls{cfggp} is more sensitive to parameter tuning than grammar design.

Hemberg et al. \cite{Hemberg2008PreIP} compared \glspl{ge} with a depth-first mapping mechanism that uses three grammars: infix (standard \gls{ge}), prefix and postfix. The results showed that different grammars can improve performance, although the authors report no significant differences.

\gls{ge2} uses two grammars, the universal and the solution grammar. The universal grammar describes the rules to construct the solution grammar. The rules are used to map the individuals and can evolve towards biasing the search space. Results showed that the evolved grammars presented some bias towards some non-terminal symbols.

Manzoni et al. \cite{Manzoni_2020} showed theoretically that different grammars of equal quality impact the performance of (1+1)-\gls{ea}. The structure of the grammar is problem dependent but can favor the search. A mutation operator that modifies the probability of selecting the grammar rules was also proposed.

\section{Adaptive Facilitated Mutation}
\label{sec:mutation_levels}
In biology, organisms have evolved to canalize the rate and effect of mutations on the phenotype. 
Gerhart et al. \cite{gerhart2007} propose that organisms adapt to new environments through regulatory changes that enable or disable pre-existing conserved components. This variation increases the probability of viable genetic mutation since core components remain unaffected by these regulatory changes. 
In sum, the modularity, adaptability, and compartmentation of genetic material in organisms allow facilitated variation through regulatory change.

Facilitated Mutation \cite{stefano2023} (FM) is a biologically inspired mutation mechanism that aims to replicate the benefits of facilitated variation.
This mechanism leverages the grammar's inherent compartmentation to regulate the mutation's frequency and destructiveness. With this mutation, each non-terminal has a different mutation probability, rather than the single mutation probability traditionally used by grammar-based \gls{ea}.

\begin{figure}[h]
\centering
\includegraphics[width=0.7\textwidth]{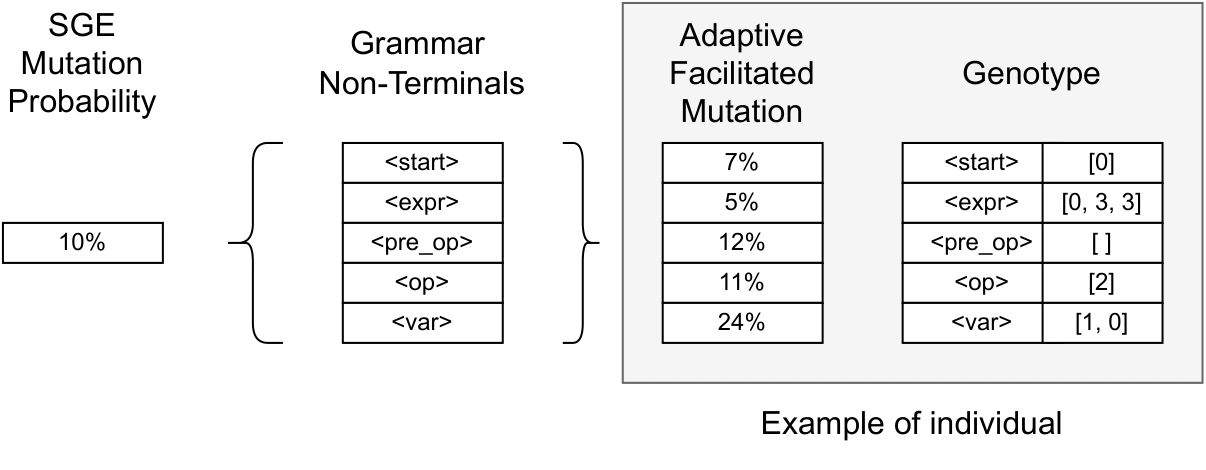}
\caption{In \gls{sge}, all genes are mutated based on a single probability (pictured left). Adaptive Facilitated Mutation uses a mutation array for each individual containing a probability for each non-terminal (pictured right).} \label{fig:mutation_array}
\end{figure}

In this work, we propose \gls{afm}, an extension of FM using an adaptive mutation array that removes the need to set a mutation probability for each non-terminal manually.
Each individual carries a \textit{mutation array} containing these mutation probabilities, illustrated in Figure~\ref{fig:mutation_array}.
For each individual in the initial population, the mutation array is initialized using a specified \textit{starting mutation probability}.

\begin{figure}[h]
\centering
\includegraphics[width=0.7\textwidth]{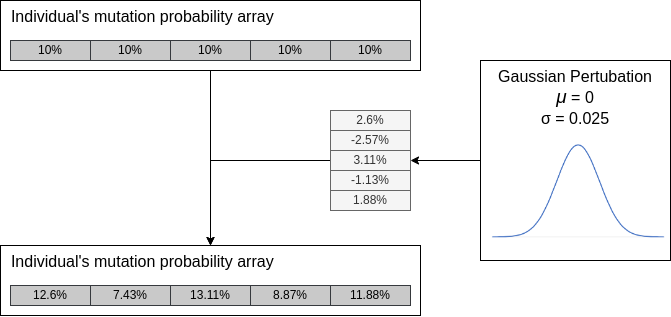}
\caption{Example of the first perturbation to the mutation probability array of an individual. This array is subsequently updated using a value sampled from a Gaussian distribution. This distribution is centered at 0 using a configurable standard deviation $\sigma$.} \label{fig:mutation_evolution}
\end{figure}

In each generation, all individual's probabilities are adjusted using a random value sampled from a Gaussian distribution with $N(0,\sigma)$, where $\sigma$ is a configurable parameter. 
Figure \ref{fig:mutation_evolution} illustrates the evolution of the mutation array.

\gls{afm} only complements mutation operators by refining the frequency and impact of mutation.
Once the mechanism determines which non-terminals to mutate, other operators should be used to alter the genotype within the defined scope.
During crossover, the offspring individual inherits the mutation array from its fittest parent.
More sophisticated inheritance mechanisms may improve this approach further, but we opted for a simple strategy to validate the approach.
In this work, we use \gls{afm} for \gls{sge} and its variants, but this method is compatible with any grammar-based \gls{gp} algorithms where it is possible to tie each codon to a corresponding non-terminal.

\subsection{Grammar Design For Adaptive Facilitated Mutation}
Since \gls{afm} leverages grammar structure, a purposefully designed grammar may enhance the method's performance.
We hypothesize that \gls{afm} is more effective in grammars with multiple non-terminals containing related symbols.
Additionally, a larger number of non-terminals may improve performance by enabling finer turning of mutation probabilities through a more detailed mutation array.
Non-terminals commonly group symbols based on semantic similarity.
For example, a grammar may use a non-terminal for all operators with a single expansion combining operators related to trigonometry (i.e., $sin$, $cos$) and the power function (i.e., $square$, $sqrt$).
These same symbols can be grouped into several non-terminals based on function rather than semantics.
Following this reasoning, trigonometric operations would be grouped in a specific non-terminal and the power function in a separate one.
In Figure \ref{fig:standard_vs_extended_grammar}, we illustrate how a grammar can be extended into a Function Grouped grammar.

\begin{figure}
\begin{subfigure}{0.5\textwidth}
    \begin{align*}
        {<}\text{start}{>}::= & {<}\text{expr}{>}\\
        {<}\text{expr}{>}::= & {<}\text{expr}{>}{<}\text{op}{>}{<}\text{expr}{>}\, |\\ 
        & {<}\text{pre\_op}{>}{(}{<}\text{expr}{>}{)}\, | \\
        & {<}\text{var}{>} \\
        {<}\text{op}{>}::= & {+} | {-} | {*} | {/} \\ 
        {<}\text{pre\_op}{>}::= & {sin} | {cos} | {sqrt} | {square}  \\ 
        {<}\text{var}{>}::= & {1.0} | {x[n]}
    \end{align*}
    \caption{Initial Grammar.}
    \label{fig:standard_example}
\end{subfigure}
\begin{subfigure}{0.5\textwidth}
\centering
    \begin{small}
    \begin{align*}
        {<}\text{start}{>}::= & {<}\text{expr\_var}{>}\\
        {<}\text{expr\_var}{>}::= &  {<}\text{expr}{>} | {<}\text{var}{>}\\ 
        {<}\text{expr}{>}::= &  {<}\text{expr\_var}{>}{<}\text{op}{>}{<}\text{expr\_var}{>}\, | \\
        & {<}\text{pre\_op}{>}{(}{<}\text{expr\_var}{>}{)}\\ 
        {<}\text{op}{>}::= & {+} | {-} | {*} | {/} \\ 
        {<}\text{pre\_op}{>}::= & {<}\text{trig\_op}{>} | {<}\text{pow\_op}{>} \\ 
        {<}\text{trig\_op}{>}::= & {sin} | {cos} \\ 
        {<}\text{pow\_op}{>}::= & {sqrt} | {square} \\ 
        {<}\text{var}{>}::= & {1.0} | {x[n]}
    \end{align*}
    \end{small}
    \caption{Function Grouped Grammar.}
    \label{fig:extended_example}
\end{subfigure}
\caption{A grammar of semantically grouped non-terminals (\ref{fig:standard_example}) can be re-constructed based on functional groups (\ref{fig:extended_example}). This procedure extends the grammar and possibly improves performance when using Adaptive Facilitated Mutation.} \label{fig:standard_vs_extended_grammar}
\end{figure}



\section{Experimental Setup}
\label{sec:experimental_setup}
We use \gls{psge} \cite{Megane2022} in all experiments as it has equal or superior performance to \gls{sge} in the selected tasks.
We compare the algorithm using Standard Mutation (SM) and Facilitated Mutation (FM).
Additionally, we compare the Standard Grammar (SG) and the Function Grouped Grammar (FG) that follows the principles outlined in Section \ref{sec:mutation_levels} (complete grammar shown in Figure \ref{fig:extended_grammar}).

We evaluate the performance of our method in three popular symbolic regression \gls{gp} benchmarks, the Quartic polynomial, the Pagie polynomial, and Boston Housing \cite{White2012,McDermott2012}.
The Quartic polynomial is defined by the mathematical expression shown in Equation \ref{eq:quartic}. The function is sampled in the interval [-1, 1] with a step of 0.1.
\begin{equation}
    x[0]^{4} + x[0]^{3} + x[0]^{2} + x[0]
    \label{eq:quartic}
\end{equation}

The Pagie polynomial is known to be a more difficult symbolic regression benchmark (Equation \ref{eq:pagie}. The outputs are computed in the interval: $-5 \leq x[0], x[1] \leq 5.4$, with step size $0.4$.
In Pagie, both features fall in the same range of values. 

\begin{equation}
    \frac{1}{1 +  x[0]^{-4} } + \frac{1}{1 + x[1]^{-4}}
    \label{eq:pagie}
\end{equation}

\begin{figure}[h!]
    \begin{small}
    \begin{align*}
        {<}\text{start}{>}::= & {<}\text{expr\_var}{>}\\
        {<}\text{expr\_var}{>}::= &  {<}\text{expr}{>} | {<}\text{var}{>}\\ 
        {<}\text{expr}{>}::= &  {<}\text{expr\_op}{>} | \\
        & {<}\text{pre\_op}{>}{(}{<}\text{expr\_var}{>}{)}\\ 
        {<}\text{expr\_op}{>}::= & {<}\text{expr}{>}{<}\text{op}{>}{<}\text{expr}{>}\, |\\ 
        & {(}{<}\text{expr}{>}{<}\text{op}{>}{<}\text{expr}{>}{)}\, \\
        {<}\text{op}{>}::= & {+} | {-} | {*} | {/} \\ 
        {<}\text{pre\_op}{>}::= & {<}\text{trig\_op}{>}\\ 
        & | {<}\text{exp\_log\_op}{>}  | {inv} \\ 
        {<}\text{trig\_op}{>}::= & {sin} | {cos} \\ 
        {<}\text{exp\_log\_op}{>}::= & {exp} | {log} \\ 
        {<}\text{var}{>}::= & {1.0} | {x[n]}
    \end{align*}
    \end{small}
\caption{Function Grouped Grammar used in the experiments. Number $x[n]$ terminals change to match the problem: 1 for Quartic, 2 for Pagie, 12 for Boston Housing} 
\label{fig:extended_grammar}
\end{figure}

The third benchmark is Boston Housing \cite{harrison1978hedonic}. This is a predictive modeling problem, where one needs to build a model to predict the price of Boston houses based on 13 features.
There are 506 instances split into 90\% for training and 10\% for test. The 13 features that compose the dataset are heterogeneous regarding their intervals, with ranges varying from $0 \leq x[3] \leq 1$ to $0.32 \leq x[11] \leq 396.9$.
This diversity likely reduces the effectiveness of our approach as all features are grouped in the $<var>$ non-terminal, making it difficult for the algorithm to distinguish between different types of features with different mutation rates.
It is possible to adjust the grammar using expert knowledge, separating the variables into non-terminals based on orders of magnitude or function.
This type of grammar design, while possibly effective, is outside the scope of this work.

The fitness functions used to evaluate the individuals consider the minimization of the \gls{rrse} between the individual's solution and the target on a data set.

Table \ref{tab:parameters} summarizes the parameters used in the experiments.
All problems use the same population size, mutation and crossover rates, tournament size, max depth, and number of generations.
In preliminary experimentation, we trialed four parameters for the 
\gls{afm}'s Gaussian perturbation: $\sigma = [0.001, 0.0025, 0.005, 0.01]$
We found that facilitated mutations' $\sigma$ parameter benefited from tuning when moved to different tasks.
Consequently, each experiment uses the best $\sigma$ value for the corresponding problem.
We repeat all experiments 100 times to investigate statistically meaningful differences between the approaches.

\begin{table}[h!]
\centering
\caption{Parameters used in experiments for Quartic, Pagie, and Boston Housing.}
\label{tab:parameters}
\begin{tabular}{|c|ccc|}
\hline
\textbf{Parameters}         & \multicolumn{1}{c|}{Quartic} & \multicolumn{1}{c|}{Pagie}  & \begin{tabular}[c]{@{}c@{}}Boston \\ Housing\end{tabular} \\ \hline
Population Size       & \multicolumn{3}{c|}{1000} \\ \hline
Generations           & \multicolumn{3}{c|}{100}  \\ \hline
Elitism               & \multicolumn{3}{c|}{10\%}  \\ \hline
Mutation            & \multicolumn{3}{c|}{Gaussian N(0,0.5)}  \\ \hline
Mutation Probability  & \multicolumn{3}{c|}{10\%} \\ \hline
Adaptive Facilitated Mutation $\sigma$ & \multicolumn{2}{c|}{0.0025}  & \multicolumn{1}{c|}{0.001}                                                      \\ \hline
Crossover Probability & \multicolumn{3}{c|}{90\%} \\ \hline
Tournament Size       & \multicolumn{3}{c|}{3}    \\ \hline
Max Depth             & \multicolumn{3}{c|}{10}   \\ \hline
\end{tabular}%
\end{table}

\section{Results}
\label{sec:results}
This section presents the results obtained for each problem in terms of the \gls{mbf} of 100 repetitions.
We statistically compare the different approaches using the Mann-Whitney test with Bonferroni correction with a significance level $\alpha = 0.05$.

\begin{figure}[h]
\centering
\includegraphics[width=0.67\textwidth]{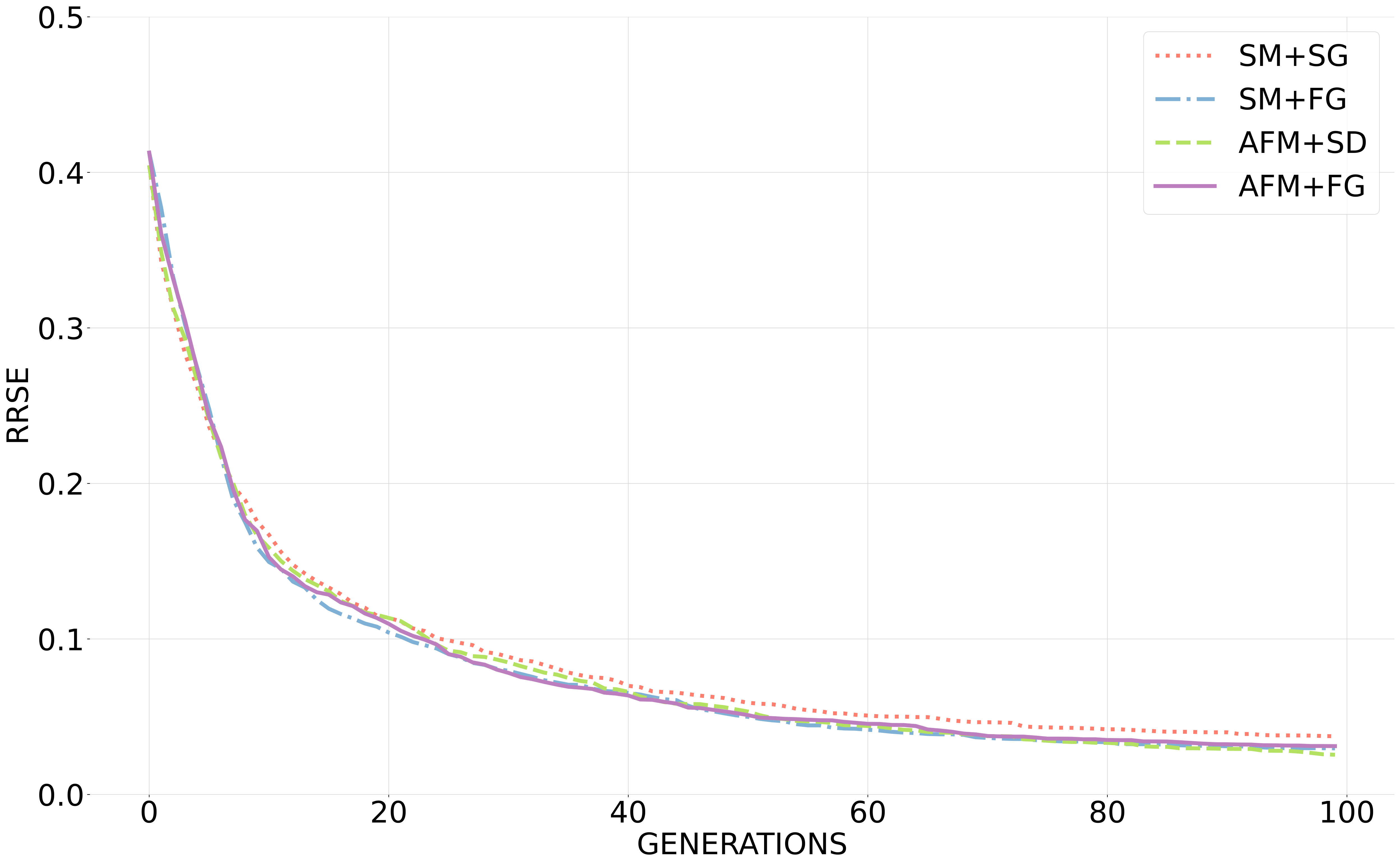}
\caption{Plot shows the mean best fitness of 100 runs for the Quartic polynomial.} \label{fig:results_quartic}
\end{figure}
Figure \ref{fig:results_quartic} shows the results for the Quartic polynomial. 
At the end of evolution, all approaches achieve a similar result in this problem, except for SM+SG, which performs significantly worse than the others (see Table \ref{tab:stats_quartic} for statistical analysis).
Quartic likely has a local optimum that evolution cannot escape without our proposed mechanism.
While SM+FG and AFM+SG statistically outperform the baseline, AFM+FG's advantages are only empirical.
We hypothesize that AFM+FG would join the other two methods with additional repetitions, creating two tiers of solution quality in Quartic.

\begin{table}[h]
\centering
\caption{P-value for Mann-Whitney Statistical Tests using Bonferroni correction with significance level $\alpha = 0.05$ for Quartic polynomial. Bold indicates that the method in the corresponding row is statistically superior.}
\label{tab:stats_quartic}
\begin{tabular}{l|llll}
\begin{tabular}[c]{@{}l@{}}\textbf{Quartic}\\ \textbf{polynomial}\end{tabular} & SM+SG                    & SM+FG                    & AFM+SG                    & AFM+FG                    \\ \hline
SM+SG                                                  & \cellcolor[HTML]{C0C0C0} & \cellcolor[HTML]{C0C0C0} & \cellcolor[HTML]{C0C0C0} & \cellcolor[HTML]{C0C0C0} \\
SM+FG                                                  &                \textbf{0.048}          & \cellcolor[HTML]{C0C0C0} & \cellcolor[HTML]{C0C0C0} & \cellcolor[HTML]{C0C0C0} \\
AFM+SG                                                  &                \textbf{0.004}          &                  0.356        & \cellcolor[HTML]{C0C0C0} & \cellcolor[HTML]{C0C0C0} \\
AFM+FG                                                  &               0.072       &                     0.760     &           0.332            & \cellcolor[HTML]{C0C0C0}
\end{tabular}
\end{table}

\begin{figure}[h!]
\centering
\includegraphics[width=0.67\textwidth]{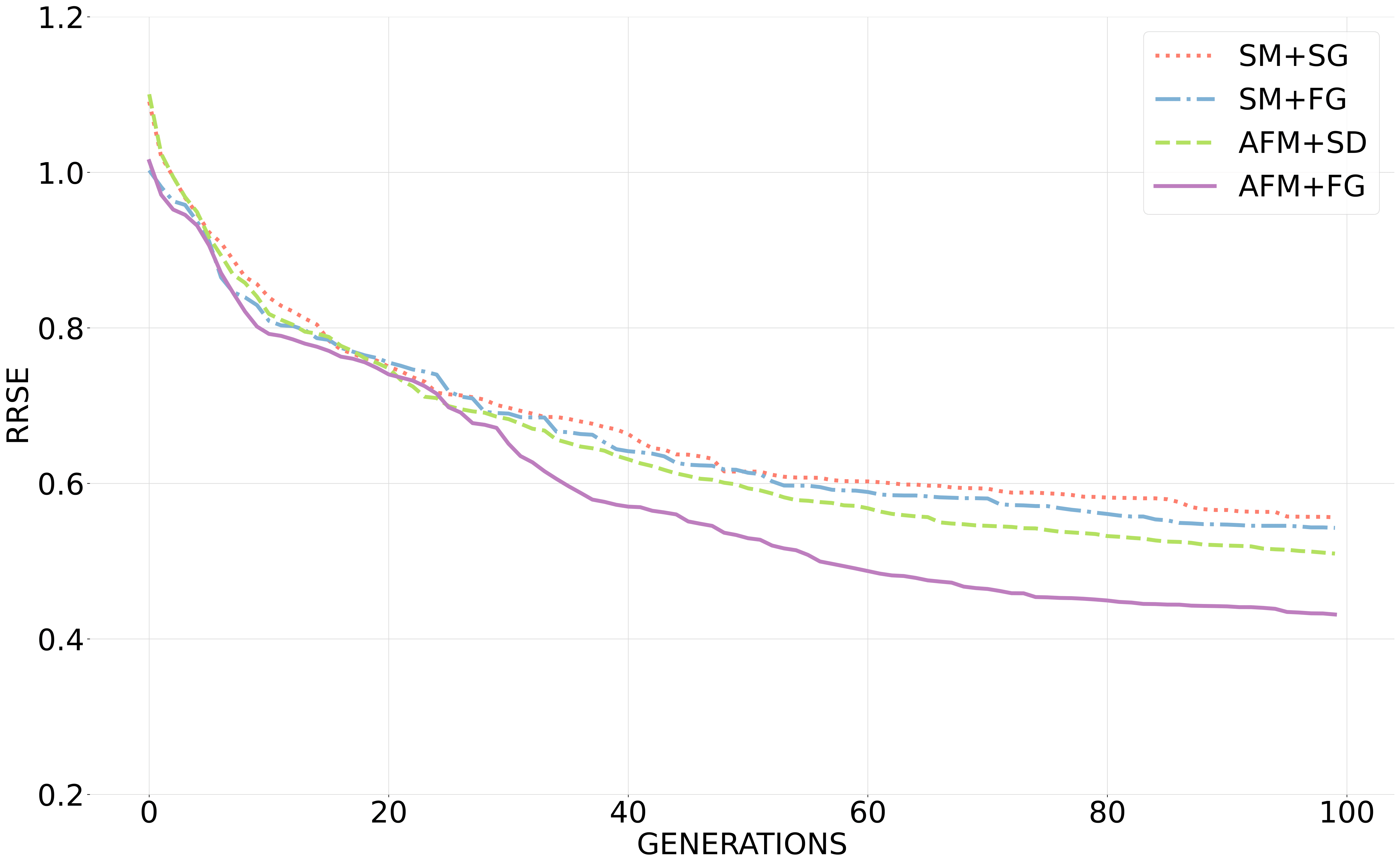}
\caption{Plot shows the mean best fitness of 100 runs for the Pagie polynomial.} \label{fig:results_pagie}
\end{figure}

Figure \ref{fig:results_pagie} shows the \gls{mbf} for the Pagie polynomial across 100 generations. Looking at the results, one can see that the advantages of the proposed approaches are clear for this problem, particularly the FG grammar.
From generation 25 until termination, both AFM solutions have a better \gls{mbf} than their SM counterparts.
While AFM is empirically superior to SM, FG amplifies the benefits of this approach.

The statistical analysis results (shown in Table \ref{tab:stats_pagie}) reveal that AFM+FG outperforms all SG approaches while SM+FG is similar.
These results suggest that our proposed grammar may be marginally superior, but considerable benefits come from combining it with the appropriate mutation.

\begin{table}[h!]
\centering
\caption{P-value for Mann-Whitney Statistical Tests using Bonferroni correction with significance level $\alpha = 0.05$ for Pagie polynomial. Bold indicates that the method in the corresponding row is statistically superior.}
\label{tab:stats_pagie}
\begin{tabular}{l|llll}
\begin{tabular}[c]{@{}l@{}}\textbf{Pagie}\\\textbf{polynomial}\end{tabular} & SM+SG                    & SM+FG                    & AFM+SG                    & AFM+FG                    \\ \hline
SM+SG                                                  & \cellcolor[HTML]{C0C0C0} & \cellcolor[HTML]{C0C0C0} & \cellcolor[HTML]{C0C0C0} & \cellcolor[HTML]{C0C0C0} \\
SM+FG                                                  &                 0.965         & \cellcolor[HTML]{C0C0C0} & \cellcolor[HTML]{C0C0C0} & \cellcolor[HTML]{C0C0C0} \\
AFM+SG                                                  &         0.426                 &                   0.400       & \cellcolor[HTML]{C0C0C0} & \cellcolor[HTML]{C0C0C0} \\
AFM+FG                                                  &             \textbf{ 0.002}            &            \textbf{ 0.001}             &           \textbf{ 0.001}              & \cellcolor[HTML]{C0C0C0}
\end{tabular}
\end{table}


\begin{figure}[h!]
\centering
    \begin{subfigure}{0.45\textwidth}
        \centering
        \includegraphics[height=4.2cm]{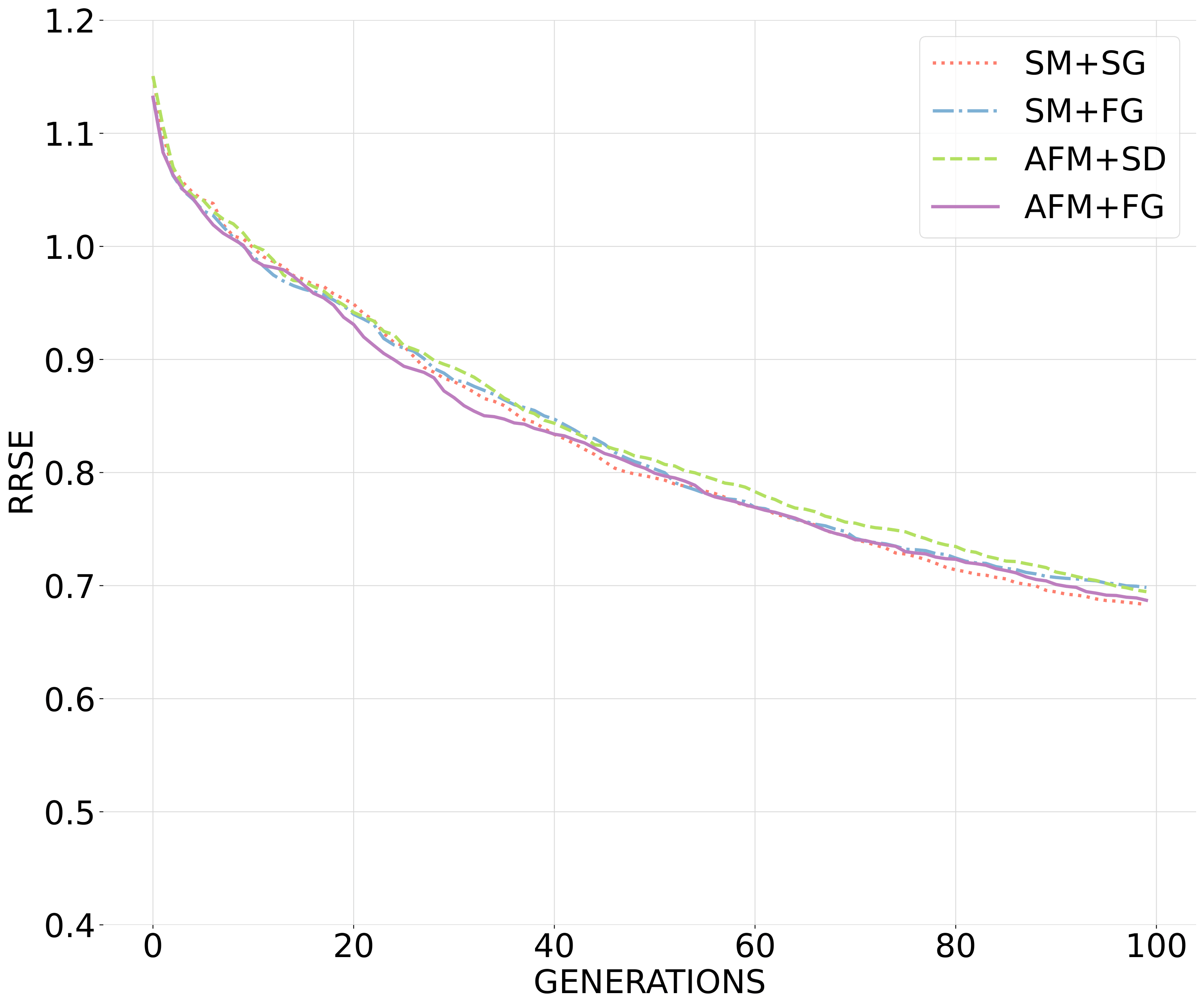}
        \label{fig:results_bh_train}
        \caption{Training}
    \end{subfigure}
    \begin{subfigure}{0.45\textwidth}
        \centering
        \includegraphics[height=4.2cm]{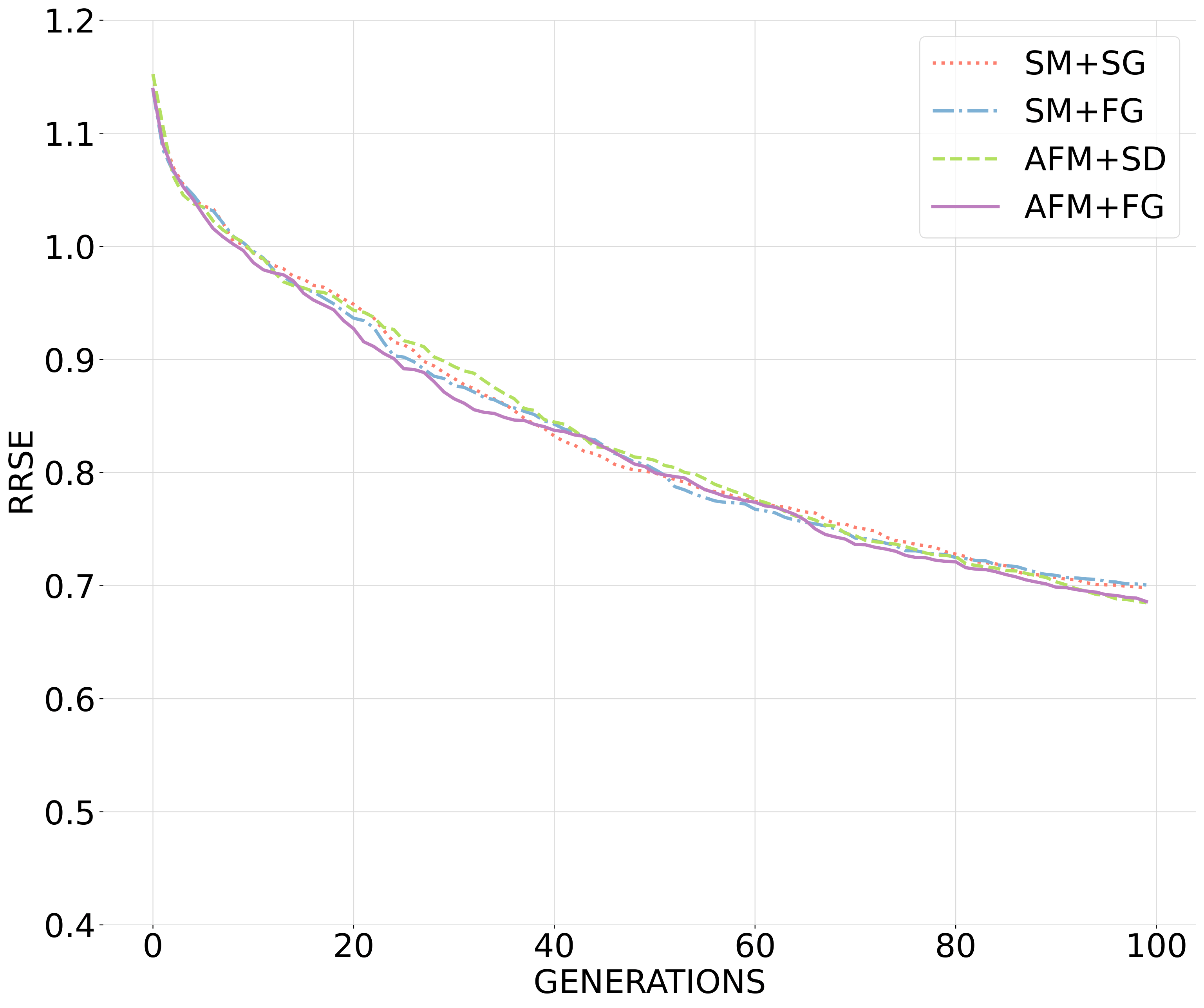}
        \label{fig:results_bh_test}
        \caption{Test}
    \end{subfigure}
\caption{Plot shows the mean best fitness of 100 runs for the Boston Housing dataset.} \label{fig:results_bh}
\end{figure}

Finally, in Figure \ref{fig:results_bh} we show the \gls{mbf} for the Boston Housing Training and Test.
A brief perusal of the results indicates that AFM+FG achieves the best \gls{mbf} in Test.
When comparing the mutations, \gls{afm} generalizes better as it maintains a similar performance between train and test.

Note that SM+SG appears to be worst at generalizing to the test data.
This is most evident when comparing SM+SG with both AFM approaches, as the differences between these methods are noticeably larger when moved to test data.
Despite the highlighted differences, statistical tests for Boston Housing Training and Test reveal no statistical differences (Tables \ref{tab:stats_bh_train} and \ref{tab:stats_bh_test}).
Given that FG cannot account for the feature variety of Boston Housing, it is remarkable that the approach still achieves competitive results, especially in the test data.
It is possible that a specifically designed FG that uses expert knowledge to group the features based on function could achieve even better and statistically significant results.

\begin{table}[h!]
\centering
\caption{P-value for Mann-Whitney Statistical Tests using Bonferroni correction with significance level $\alpha = 0.05$ for Boston Housing Training. Bold indicates that the method in the corresponding row is statistically superior.}
\label{tab:stats_bh_train}
\begin{tabular}{l|llll}
\begin{tabular}[c]{@{}l@{}}\textbf{Boston Housing}\\\textbf{Training}\end{tabular} & SM+SG                    & SM+FG                    & AFM+SG                    & AFM+FG                    \\ \hline
SM+SG                                                  & \cellcolor[HTML]{C0C0C0} & \cellcolor[HTML]{C0C0C0} & \cellcolor[HTML]{C0C0C0} & \cellcolor[HTML]{C0C0C0} \\
SM+FG                                                  &               0.530           & \cellcolor[HTML]{C0C0C0} & \cellcolor[HTML]{C0C0C0} & \cellcolor[HTML]{C0C0C0} \\
AFM+SG                                                  &                 0.263         &          0.763                & \cellcolor[HTML]{C0C0C0} & \cellcolor[HTML]{C0C0C0} \\
AFM+FG                                                  &                 0.531         &       0.361                   &          0.548             & \cellcolor[HTML]{C0C0C0}
\end{tabular}
\end{table}

\begin{table}[h!]
\centering
\caption{P-value for Mann-Whitney Statistical Tests using Bonferroni correction with significance level $\alpha = 0.05$ for Boston Housing Test. Bold indicates that the method in the corresponding row is statistically superior.}
\label{tab:stats_bh_test}
\begin{tabular}{l|llll}
\begin{tabular}[c]{@{}l@{}}\textbf{Boston Housing}\\\textbf{Test}\end{tabular} & SM+SG                    & SM+FG                    & AFM+SG                    & AFM+FG                    \\ \hline
SM+SG                                                  & \cellcolor[HTML]{C0C0C0} & \cellcolor[HTML]{C0C0C0} & \cellcolor[HTML]{C0C0C0} & \cellcolor[HTML]{C0C0C0} \\
SM+FG                                                  &           0.929               & \cellcolor[HTML]{C0C0C0} & \cellcolor[HTML]{C0C0C0} & \cellcolor[HTML]{C0C0C0} \\
AFM+SG                                                  &              0.377            &        0.333                 & \cellcolor[HTML]{C0C0C0} & \cellcolor[HTML]{C0C0C0} \\
AFM+FG                                                  &             0.364             & 0.403                   &        0.963                  & \cellcolor[HTML]{C0C0C0}
\end{tabular}
\end{table}

\section{Conclusion}
\label{sec:conc}
In this paper, we propose \gls{afm}, a mutation method that leverages grammar-based \gls{gp}'s properties to replicate natural evolutionary phenomena.
This approach divides the single mutation probability commonly found in such approaches into a mutation array, where each grammar non-terminal has a corresponding mutation rate. Each individual has a mutation array that co-evolves with the genetic code. A randomly sampled value from a Gaussian distribution adjusts the mutation rates of all individuals in each generation.

We also propose a grammar-design approach, Function Grouped Grammars, to enhance the effectiveness of the mutation proposed.
Function Grouped Grammars organize non-terminals based on functional similarity rather than the semantic similarity common in the field.
We compare our proposals with a baseline (standard mutation and standard grammar) and find that, when combined, our approaches are statistically superior or similar to the baseline in three relevant \gls{gp} benchmarks.

This approach still requires parameter tuning, one of the problems tackled by the literature by proposing adaptive mutations. However, few of these approaches consider different values for different symbols \cite{Coelho2016}. 
This work shows that an adaptive mutation rate can be beneficial for search and grammar design amplifies these benefits.

\subsection{Future Work}
The results of our experiments are promising, but additional tests are essential to validate our approach further.
The benchmarks addressed are relevant in \gls{gp}, but a more extensive (and varied) set of benchmarks could bring meaningful insights into the general applicability of the FG+FM.

Regarding \gls{afm}, we use a fixed starting mutation rate in all experiments.
Our method leverages adaptability as a tool for improved evolution, but we did not investigate the potential of \gls{afm} as a replacement for mutation rate tuning.
Further development of \gls{afm} may also lead to an implementation that does not rely on Gaussian distributions and $sigma$ tuning for perturbations. 
In the future, we want to explore alternatives where \gls{afm} is a competitive, parameter-less alternative to standard mutation.
Another line of work is to experiment with different inheritance mechanisms during crossover. 
In this work, the offspring inherited the array of probabilities from the most fitted parent.
It would be interesting to explore the random selection of the parent that passes the array or the application of the existing SGE crossover to the mutation arrays of the parents.
Such approaches would better preserve the advantages of co-evolution, possibly improving results. 

Function Grouped Grammars can also be further investigated.
In this work, we still apply this idea conservatively.
Considering the FG used in experiments, it is still possible to separate constants from variables and commutative operators from non-commutative operators.
Barring small details, we use the same grammar for all tasks.
Applying the same approach to problems requiring more complex grammar would be interesting, as such problems yield more opportunities for function grouping.

Finally, the applicability of these ideas to different, compatible grammar-based \gls{gp} must also be investigated.
While this works focuses on \gls{psge}, the same approach is easily applicable in \gls{sge} \cite{Loureno2018}, \gls{copsge} \cite{Megane2022gecco}, and even more different systems like $\pi$GE \cite{ONeill2004}.
Any grammar-based approach where genes are tied to a non-terminal may benefit from FG+FM.




%

\section*{Acknowledgments}
This work was funded by FEDER funds through the Operational Programme Competitiveness Factors - COMPETE and national funds by FCT - Foundation for Science and Technology (POCI-01-0145-FEDER-029297, CISUC - UID/CEC/00326/2020) and within the scope of the project A4A: Audiology for All (CENTRO-01-0247-FEDER-047083) financed by the Operational Program for Competitiveness and Internationalisation of PORTUGAL 2020 through the European Regional Development Fund. The first author is funded by FCT, Portugal, under the grant UI/BD/151053/2021 and the second under the grant 2022.10174.BD.

%
%
%
\bibliographystyle{splncs04}
\bibliography{bibliography}




\end{document}